\documentclass[11pt]{article}
\usepackage{xr-hyper}
\usepackage[preprint]{acl}

\usepackage{times}
\usepackage{latexsym}
\usepackage[T1]{fontenc}
\usepackage[utf8]{inputenc}
\usepackage{microtype}
\usepackage{inconsolata}
\usepackage{graphicx}

\usepackage{booktabs,multirow,array,tabularx}
\usepackage[table]{xcolor}
\definecolor{heat}{HTML}{2E7D32}
\definecolor{sub}{HTML}{7A7A76}
\newcolumntype{R}{>{\raggedleft\arraybackslash}X}
\newcolumntype{G}{>{\raggedleft\arraybackslash\leavevmode\color{sub}}X}

\usepackage{booktabs}
\usepackage{multirow}
\usepackage{amsmath}
\usepackage{placeins}

\title{Multi-Agent Memory Inconsistency:\\
Where Factual Conflicts Disappear in LLM Agents}
\author{
  \textbf{Hejia Zhang\textsuperscript{1,3}}\thanks{This work was done during an internship at Intel.},
  \textbf{Zhongming Yu\textsuperscript{1}},
  \textbf{Yimin Ding\textsuperscript{1}},
  \textbf{Wei-Ting Chen\textsuperscript{1}},
\\
  \textbf{Hengjia Yu\textsuperscript{1}},
  \textbf{Wentao Ni\textsuperscript{1}},
  \textbf{Yijun Liu\textsuperscript{1}},
  \textbf{Cong Xu\textsuperscript{2}},
\\
  \textbf{Xin Chen\textsuperscript{3}},
  \textbf{Matrix Yao\textsuperscript{3}},
  \textbf{Fan Zhao\textsuperscript{3}},
  \textbf{Jishen Zhao\textsuperscript{1}}
\\
\\
  \textsuperscript{1}UC San Diego,
  \textsuperscript{2}Hewlett Packard Enterprise,
  \textsuperscript{3}Intel
}

\begin{document}
\maketitle

\begin{abstract}
Multi-agent delegation lets LLM systems distribute evidence collection across workers with separate contexts.
This creates a consistency challenge: facts retrieved by workers may be missing from the evidence available to the lead for a joint conclusion.
We introduce a benchmark and evidence-flow profiler for studying multi-agent memory inconsistency through controlled factual conflicts in Harvey LAB document-review tasks.
Pooled across three models with comparable generation spending, requiring delegation increases retrieval of both conflicting claims from 50.2\% to 68.7\% relative to single-agent execution, yet conflict detection falls from 16.2\% to 13.2\%.
Within this setting, workers report both claims for only 35.2\% of retrieved pairs, and pooled matched comparisons show higher detection than single-agent execution when one worker reads both claims but lower detection when the claims are split across workers.
Worker-to-lead handoffs are a central target for improving conflict detection in multi-agent LLMs.

\end{abstract}

\begin{figure}[!t]
  \centering
  % Edit figures/source/motivation-tidy.svg and run scripts/export_figures.py 1.
  \includegraphics[width=\columnwidth]{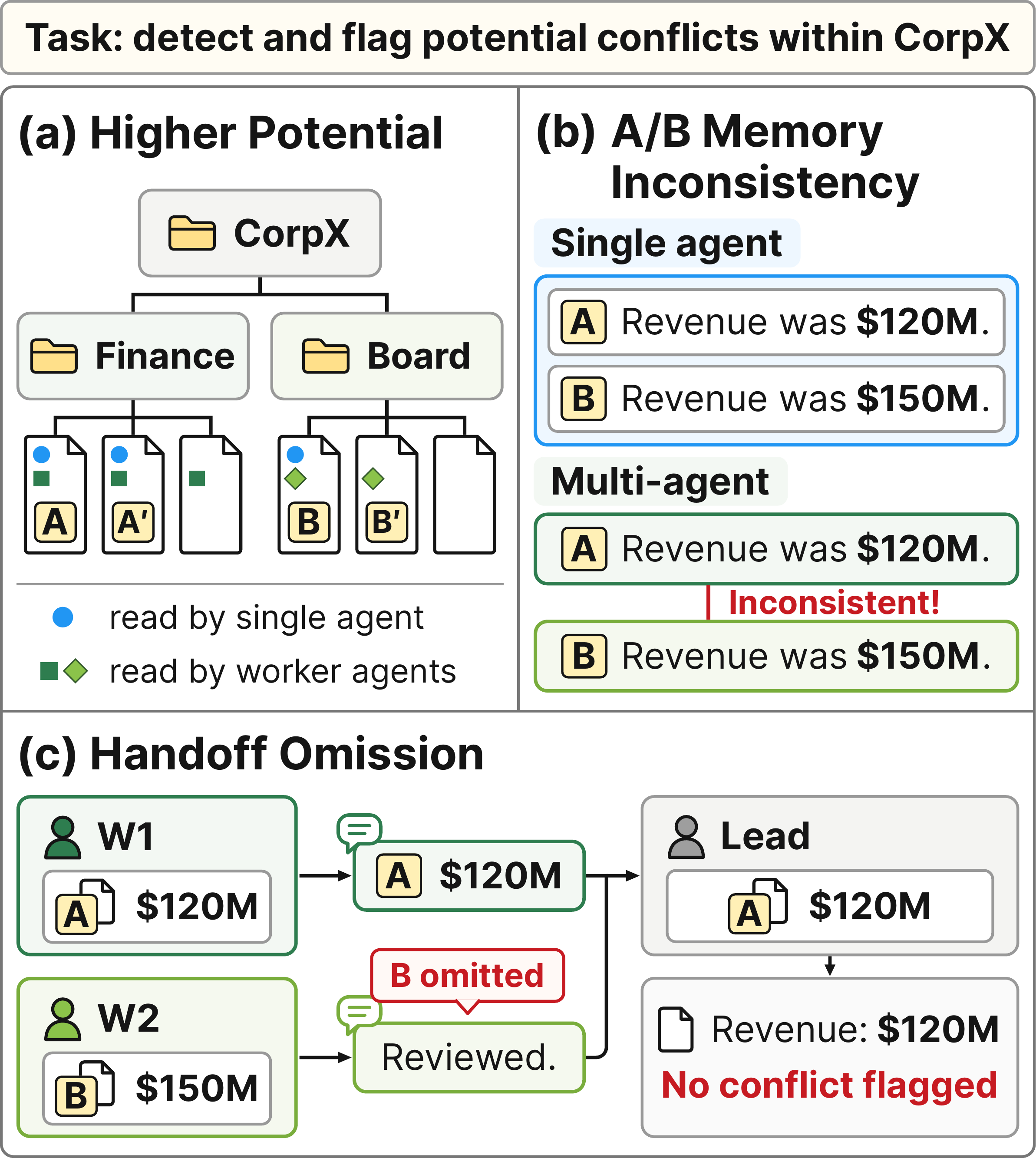}
  \caption{Multi-agent memory inconsistency. A and B give conflicting
  CorpX FY2025 revenue figures.
  (a) The single agent retrieves A, B, and A$'$; two workers also retrieve B$'$, completing
  a second fact pair.
  (b) Focusing on A/B, both facts occupy one single-agent context or separate worker contexts.
  (c) B is read but omitted from a worker summary, leaving the lead with only A
  and the conflict undetected.}
  \label{fig:motivation}
\end{figure}

\section{Introduction}
\label{sec:introduction}

Multi-agent LLM systems scale complex workflows by distributing work across agents with separate contexts.
In lead--worker \textbf{(NOT SURE the definition is widely used. See my comment)} architectures, a lead assigns subtasks to specialized workers and combines their findings \citep{fourney2024magentic}.
%  XIN CHEN COMMENT I check the reference, there is not a clear definition, lead--worker. According to https://www.openlayer.com/blog/multi-agent-system-architecture-guide, I perfer to orchestrator–workers. Since this paper is not to design a new archiecture, I think it is better to use existing names
Research systems use this structure to explore multiple information sources in parallel and report gains over single-agent execution \citep{anthropic2025research}.
Controlled evaluations nevertheless show that these gains depend on task structure and coordination choices \citep{kim2026scaling}.

%In single-agent systems, working memory tracks task state across successive model calls \citep{sumers2024coala}; under delegation, task information is distributed across agents' contexts, messages, and working artifacts.
%An architectural view of such distributed memory identifies consistency across local and shared state as an open challenge \citep{yu2026multiagentmemory}.
%We study \emph{multi-agent memory consistency}: whether task-relevant evidence remains available across these information states where a joint conclusion requires it.
%As Figure~\ref{fig:motivation} illustrates, a worker can accurately summarize its assigned documents yet omit a claim needed to detect a conflict with another worker's findings.

% Xin Chen comment, in figure 1 or other part, can we use multi-tread to show our ideas such as mutex aims at keeping memory constist and data right. I think the challegne of this paper is to show the reviewer our movitation is improtant ane let the understand. multithread is a good exampel to let them image the data consistatn problem in multi-agent

Delegation, despite its parallel exploration gains, reshapes agent working memory.
In single-agent systems, working memory tracks task state across successive model calls \citep{sumers2024coala}; under delegation, task information is distributed across agents' contexts, messages, and shared artifacts, creating a coordination challenge \citep{yu2026multiagentmemory}.
To analyze this gap, we study \emph{multi-agent memory consistency}: whether task-relevant evidence remains available across local and shared states, where a joint conclusion requires it.
As Figure~\ref{fig:motivation} illustrates, a worker can accurately summarize its assigned documents yet omit a claim needed to detect a conflict with another worker's findings.

% Keep the headline paragraph from opening with one line at a page bottom.
\begingroup
\clubpenalty=10000
\brokenpenalty=10000
We examine this problem through factual conflict detection in document-review tasks, comparing single-agent execution (\emph{None}), discretionary delegation (\emph{Optional}), and mandatory delegation (\emph{Required}).
Required reserves source inspection for workers assigned to document partitions.
Across three models and two runs per model under common per-task dollar allowances, Required increases retrieval of both conflicting claims (\emph{Source Reads Pair Recall}) from 50.2\% to 68.7\% relative to None, yet conflict detection (\emph{Judge Pass}) falls from 16.2\% to 13.2\%.
Optional achieves the highest pooled detection rate, at 24.2\%.
\par\endgroup

Tracing the conflicting claims leads to 
%locates 
the largest drop in evidence preservation at the worker-to-lead handoff.
Under Required, Pair Recall falls from 68.7\% at Source Reads to 24.4\% at Lead Evidence.
Workers report both claims for only 35.2\% of the pairs they collectively retrieve, leaving the lead without the complete pair its workers retrieved.

%We
As the first stage-wise empirical study of multi-agent memory inconsistency, this paper makes three key contributions:
\begin{itemize}
\item \textbf{Benchmark and profiler (\S\ref{sec:study-design}).}
We construct 73 controlled numerical conflicts across 11 Harvey LAB document-review tasks \citep{grupen2026lab}, with a profiler tracing both claims through Source Reads, Worker Reports, Lead Evidence, and the Final Report.
Pair Recall measures evidence coverage separately from semantic conflict detection under the three delegation policies.
\item \textbf{Evidence loss at the worker-to-lead handoff (\S\ref{sec:conversion}).}
Pooled across models and runs under comparable generation spending, Required retrieves more complete evidence pairs than None but detects fewer conflicts.
The largest measured loss occurs at the worker-to-lead handoff, where workers omit claims.
\item \textbf{Different outcomes for Same and Split evidence (\S\ref{sec:matched}).}
In pooled cohorts defined by Required's Source Reads, Required detects more conflicts than None on identical cases when one worker reads both claims (34.2\% versus 22.5\%), but fewer when the claims are split across workers (10.0\% versus 18.9\%).
\end{itemize}

\section{Background and Related Work}
\label{sec:related}
% Queue the two-column methods overview on page 2 so it opens page 3.
\begin{figure*}[t]
  \centering
  \includegraphics[width=\textwidth]{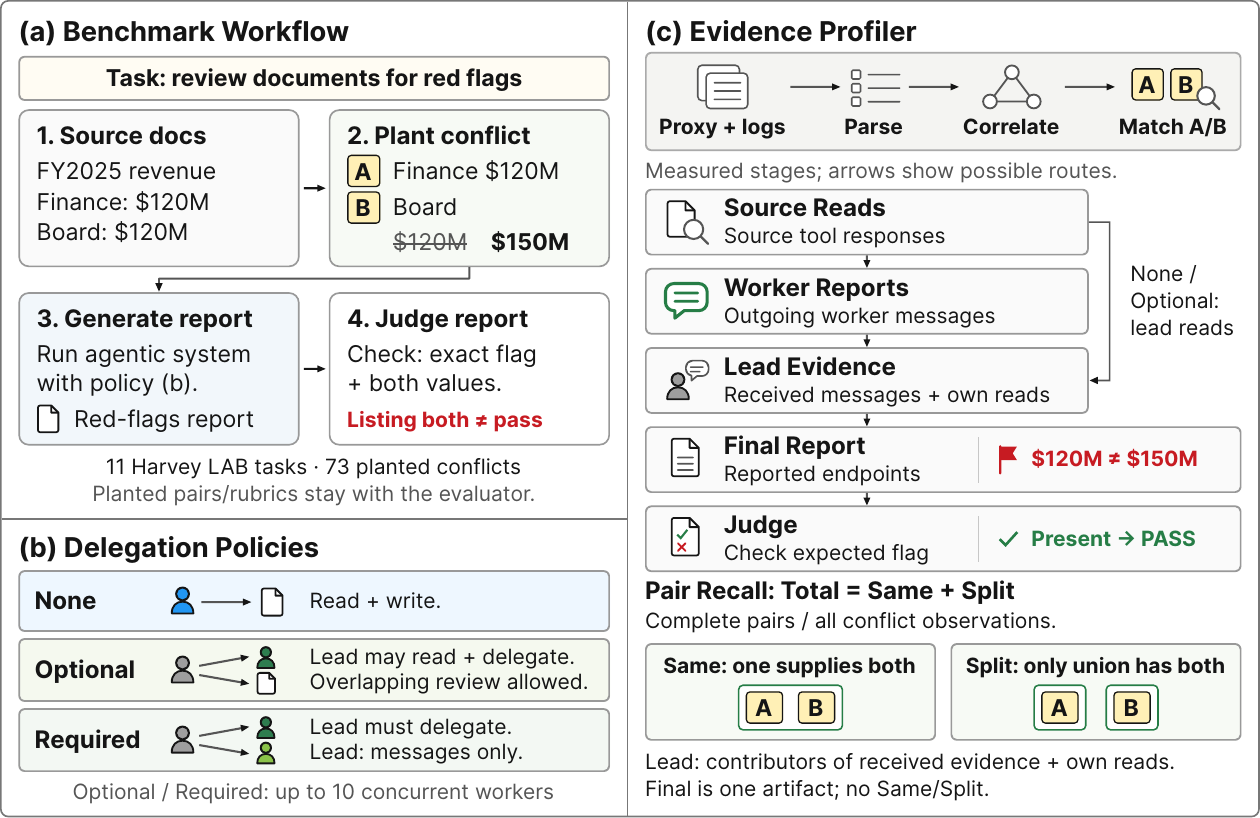}
  \caption{Benchmark construction and evidence measurement.
  (a) Controlled edits introduce numerical conflicts into Harvey LAB
  document collections; agents generate a red-flags report, which is
  evaluated against the specified conflicts.
  (b) None, Optional, and Required control delegation and lead source access.
  (c) The profiler correlates captured traffic, agent/tool events, and
  artifacts to measure Pair Recall across four stages and a separate
  Judge Pass outcome. Same/Split separates pairs supplied by one individual
  from pairs supplied only across individuals; at the lead, provenance
  refers to immediate contributors.}
  \label{fig:benchmark-metrics}
\end{figure*}

\subsection{Scaling and Delegation in Agent Systems}

LLM agents interleave reasoning traces with actions on external environments \citep{yao2023react} and retain information across steps and trials through memory and reflection \citep{shinn2023reflexion,park2023generative,wang2024survey}.
Multi-agent systems divide tasks among communicating agents with assigned roles \citep{li2023camel,qian2024chatdev,guo2024survey}.
In the orchestrator--worker pattern, a lead plans and synthesizes while workers execute subtasks \citep{anthropic2024agents,anthropic2025research}; Magentic-One similarly separates orchestration from specialized execution \citep{fourney2024magentic}.
AutoGen supports configurable conversations among agents and tools \citep{wu2024autogen}, while MetaGPT structures collaboration through specialized roles and standard operating procedures \citep{hong2024metagpt}.

Coding agents have followed the same path.
SWE-bench evaluates the resolution of real GitHub issues by editing a repository \citep{jimenez2024swebench}.
SWE-agent addresses it through a purpose-built agent--computer interface \citep{yang2024sweagent}, OpenHands provides an open platform for developer agents that supports coordination among multiple agents \citep{wang2025openhands}, and MAGIS divides planning and coding among four role-specific agents \citep{tao2024magis}.
Current coding harnesses provide subagents that work in their own context windows or threads and return summaries to the main agent \citep{anthropic2026subagents,openai2026subagents}.
For long-form writing, STORM gathers material through simulated conversations in which writers with different perspectives question a source-grounded expert \citep{shao2024storm}, and DeepResearch Bench evaluates the reports and citations of deep-research agents \citep{du2026deepresearch}.

Multi-agent debate was proposed to improve factuality and reasoning \citep{du2024debate}, yet a single agent with strong prompts nearly matches the best discussion methods \citep{wang2024rethinking}, and debate systems do not reliably outperform self-consistency or ensembling without careful tuning \citep{smit2024mad}.
The benefit of multi-agent over single-agent systems diminishes as model capabilities improve \citep{gao2025single}, and single agents match or outperform multi-agent systems on multi-hop reasoning when reasoning tokens are held constant \citep{tran2026single}.
Controlled comparisons show that agent-system performance depends on task structure \citep{kim2026scaling}.
Agentic systems can be designed automatically \citep{hu2025adas}, although automatically designed multi-agent systems can incur greater cost without outperforming strong single-agent baselines \citep{jwalapuram2026illusion}.

Context policies govern which evidence agents receive and retain \citep{yu2026codenib}; history edits can also increase execution cost when they disrupt prefix caching, even with fewer prompt tokens \citep{yu2026physicalcontext}.
For coding agents, masking old observations halves cost relative to an unmanaged history while matching the solve rate of LLM summarization \citep{lindenbauer2025complexity}, and compaction can silently remove task constraints \citep{salle2026compaction}.
Iterative context rewriting can likewise erode details over time \citep{zhang2026ace}.
For delegated workflows, we track which retrieved facts reach the lead and final report, together with conflict detection and generation spending.

\subsection{Distributed Evidence and Multi-Agent Failures}

MAST categorizes failures in multi-agent traces across frameworks \citep{cemri2025fail}, and AgentDebug diagnoses errors in agent trajectories and supplies corrective feedback \citep{zhu2025agentdebug}.
Failure attribution methods aim to identify the agent and step responsible for a failed trajectory \citep{zhang2025whichagent,zhang2026agentracer}.
HiddenBench tests collective reasoning and communication by supplying complete information to single agents and distributing facts across multi-agent groups \citep{li2026hiddenbench}.
In our setting, both single-agent and multi-agent systems must first retrieve evidence from document collections and then produce long-form reports, exposing what happens before and after workers select findings to report.

A practitioner account attributes conflicting subagent outputs to context that the subagents do not share and recommends sharing full agent traces \citep{yan2025dont}, whereas recommended subagent designs return condensed summaries to the lead \citep{anthropic2025context,anthropic2025agentsdk}.
An information-theoretic analysis treats a model that compresses context for another model as a noisy channel and finds that the mutual information between a context and its compression predicts downstream performance \citep{he2026information}.
We measure this handoff for specified claims.

% Keep a word from breaking across the page beneath Figure 2.
\begingroup
\brokenpenalty=10000
Tracing tools provide a complementary basis for inspecting these workflows.
AgentSight correlates LLM traffic with system events \citep{zheng2025agentsight}, and PROV-AGENT connects agent interactions to workflow provenance \citep{souza2025provagent}.
Our profiler uses captured execution records to track specified conflicting claims: which agents read them, which contributors report them, and whether the final report identifies their discrepancy.
\par\endgroup

\subsection{Agent Memory and Conflicting Evidence}

MemGPT manages memory tiers to extend an agent's limited context window \citep{packer2023memgpt}, and surveys organize the design and evaluation of agent memory \citep{zhang2025memorysurvey,hu2025memory}.
Benchmarks test long-term memory across conversation sessions \citep{maharana2024locomo,wu2025longmemeval}.
G-Memory organizes multi-agent collaboration trajectories in a three-tier graph hierarchy \citep{zhang2025gmemory}.
An architectural account of multi-agent memory separates update visibility from conflict resolution over shared or distributed state \citep{yu2026multiagentmemory}.
StateFuse designs memory interfaces that preserve conflicting claims \citep{volkov2026statefuse}, while Facts Without Rules traces privacy-related usage constraints lost during handoffs \citep{wang2026facts}.
Inconsistency also arises among the context views served to a single coding agent: CodeNib compiles lexical, dense, and structural views of one commit so that their results can be cross-checked, and reports that incrementally maintained views can diverge from a full rebuild even when every update completes without error \citep{yu2026codenib}.
We trace known factual comparisons through source inspection and worker reporting in document analysis, measuring whether both claims remain available for a joint conclusion.

Knowledge-conflict studies examine how models respond when supplied passages contradict parametric knowledge or one another \citep{longpre2021entity,chen2022rich,xie2024chameleon,xu2024conflicts}.
WikiContradict and ContraDoc evaluate models on contradictory passages and self-contradictory documents \citep{hou2024wikicontradict,li2024contradoc}.
Model performance also degrades when relevant information sits in the middle of a long context \citep{liu2024lost}, and summarizing the insights repeated across many documents remains an open challenge for long-context LLMs and RAG systems \citep{laban2024haystack}.
These evaluations give a single system the evidence to be reconciled, whereas our setting requires separate agents to retrieve and report both claims before any model can compare them.

FEVER evaluates supporting-evidence recovery alongside claim verification \citep{thorne2018fever}, and HotpotQA evaluates supporting facts alongside multi-hop answers \citep{yang2018hotpotqa}.
Adapting the complete-evidence-group principle to the two claims defining a planted conflict lets us measure evidence recovery at each stage separately from semantic detection.

\section{Benchmark and Measurement}
\label{sec:study-design}

We evaluate whether agents flag planted numerical conflicts in their reports, using the document edits, delegation policies, and profiling pipeline summarized in Figure~\ref{fig:benchmark-metrics}.

\subsection{Conflict Injection in Harvey LAB}
\label{sec:injection}
% Queue the results overview on page 4 so the full-width float opens page 5.
\begin{figure*}[!t]
  \centering
  \includegraphics[width=0.95\textwidth]{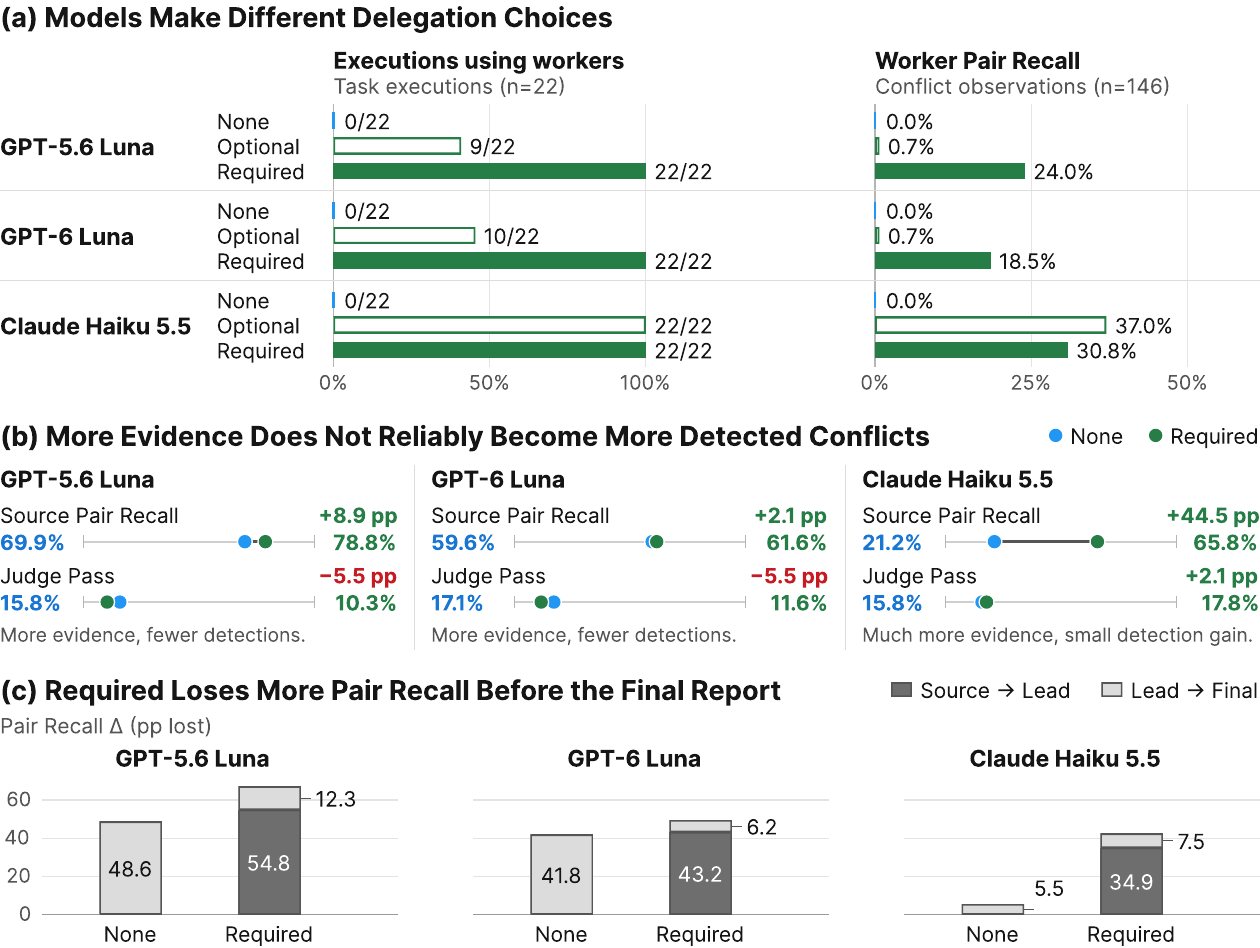}\vspace{-5pt}
  \caption{Summary of observed patterns in delegation, evidence
  preservation, and conflict detection.
  (a) Model-dependent Optional behavior motivates measuring Required
  to separate delegation choice from delegated execution.
  (b) Required's greater evidence collection translates poorly into Judge Pass.
  (c) Required has larger Pair Recall drops for all three models, with most of its
  loss between Source Reads and Lead Evidence.
  Worker Reports uses all 146 conflict observations per model/policy, including
  executions without workers; None has zero worker contribution.
  Percentage-point changes use unrounded rates.}
  \label{fig:study-insights}
\end{figure*}

We build on 11 Harvey LAB due-diligence tasks in which agents review document directories and write red-flags reports \citep{grupen2026lab}.
Each task retains its original instructions and document collection, with controlled factual edits and conflict-specific grading criteria.

To plant conflict $i$, we replace selected appearances of a recurring numerical value $A_i$ with $B_i$ in designated directories, leaving $A_i$ unchanged elsewhere.
Each conflict specification records the values, source locations, permitted equivalent representations, edits, and grading criteria.
Five to seven conflicts appear per task, totaling 73; all conflicts in a task are evaluated against the same report.

Agents receive the ordinary task prompt and edited documents.
The evaluator keeps the conflict specifications and grading rubrics separate from the agent inputs.
A successful report flags the discrepancy, states the need for reconciliation, and accurately treats the quantitative comparison.

\subsection{Delegation Policies}
\label{sec:delegation}

Every policy receives the same task and edited corpus.
Under \emph{None}, one agent inspects documents and writes the report.
Under \emph{Optional}, the lead may inspect sources itself, create workers, and assign overlapping reviews.
Under \emph{Required}, workers inspect assigned document partitions while the lead coordinates and synthesizes their findings without reading source documents.
Workers can also contribute to a shared report draft.
Optional and Required permit up to ten concurrent workers, with additional instances possible across continuation phases.

Optional measures whether models choose to delegate, whereas Required evaluates execution under mandatory delegation.
It adapts the separation of coordination and execution used by Magentic-One, which routes tool use through specialized agents \citep{fourney2024magentic}, and the original AgentOrchestra planner, which delegates environment interaction to sub-agents \citep{zhang2025agentorchestra}.
In our document tasks, this policy combines mandatory delegation, partitioned source assignments, and restricted lead access.

\subsection{Evidence-Flow Profiling}
\label{sec:profiling}

Our profiler captures model requests and responses through a local proxy and correlates them with SDK events, native traces, tool outputs, and report artifacts.
Agent and tool-call identifiers attribute evidence to its reader, sender, or recipient within the trajectory prefix that produced the graded report.
Source evidence is counted from returned content, including search excerpts.

We follow four measurement stages.
\textbf{Source Reads} captures endpoints visible in source-tool outputs; \textbf{Worker Reports} captures outgoing worker messages and completion reports, including peer-directed messages.
\textbf{Lead Evidence} combines incoming evidence with the lead's permitted source reads.
\textbf{Final Report} contains endpoints in the selected, graded deliverable.
Figures and tables abbreviate these as Source, Worker, Lead, and Final.

A uniform matcher identifies endpoints through fact-specific anchors and verified equivalent representations.
We track outgoing messages and incoming receipts separately, using plaintext collaboration traces for GPT-6 delegated executions.
Source Reads, Worker Reports, and Lead Evidence accumulate observations through the graded checkpoint; Final Report measures the selected deliverable.

\subsection{Pair Recall and Conflict Detection}
\label{sec:metrics}
% Queue the full-width table on page 5 so that it opens page 6,
% leaving page 7 for Figure 4.
% Generated by scripts/build_main_results_table.py; edit the generator.
\begin{table*}[t]
\centering
\begingroup
\small
\setlength{\tabcolsep}{3pt}
\renewcommand{\arraystretch}{1.14}
\setlength{\aboverulesep}{0pt}
\setlength{\belowrulesep}{0pt}
\setlength{\extrarowheight}{1.5pt}
\begin{tabularx}{\textwidth}{l@{\hspace{6pt}}l RGG RGG RGG rrr}
\toprule
 & & \multicolumn{3}{c}{Source Reads} & \multicolumn{3}{c}{Worker Reports} & \multicolumn{3}{c}{Lead Evidence} & \multicolumn{1}{r}{Final Report} & \multicolumn{1}{r}{Judge} & \multicolumn{1}{r}{Cost} \\
\cmidrule(lr){3-5}\cmidrule(lr){6-8}\cmidrule(lr){9-11}
Model & Delegation & Total & Same & Split & Total & Same & Split & Total & Same & Split & \multicolumn{1}{r}{Both} & \multicolumn{1}{r}{Pass} & \multicolumn{1}{r}{($\times$ budget)} \\
\midrule
\multirow{3}{*}{\shortstack[l]{GPT-5.6\\Luna}} & None & \cellcolor{heat!39}69.9 & 69.9 & 0.0 & \multicolumn{3}{c}{\color{sub}---} & \cellcolor{heat!39}\textbf{69.9} & 69.9 & 0.0 & \cellcolor{heat!12}21.2 & \cellcolor{heat!9}15.8 & 0.96 \\
 & Optional & \cellcolor{heat!38}67.1 & 65.1 & 2.1 & \cellcolor{heat!0}0.7 & 0.7 & 0.0 & \cellcolor{heat!35}63.0 & 63.0 & 0.0 & \cellcolor{heat!15}\textbf{26.0} & \cellcolor{heat!13}\textbf{23.3} & 0.95 \\
 & Required & \cellcolor{heat!44}\textbf{78.8} & 32.9 & 45.9 & \cellcolor{heat!14}\textbf{24.0} & 18.5 & 5.5 & \cellcolor{heat!14}24.0 & 18.5 & 5.5 & \cellcolor{heat!7}11.6 & \cellcolor{heat!6}10.3 & 1.02 \\
\addlinespace[5pt]
\multirow{3}{*}{\shortstack[l]{GPT-6\\Luna}} & None & \cellcolor{heat!34}59.6 & 59.6 & 0.0 & \multicolumn{3}{c}{\color{sub}---} & \cellcolor{heat!34}59.6 & 59.6 & 0.0 & \cellcolor{heat!10}17.8 & \cellcolor{heat!10}17.1 & 0.95 \\
 & Optional & \cellcolor{heat!37}\textbf{65.1} & 63.7 & 1.4 & \cellcolor{heat!0}0.7 & 0.7 & 0.0 & \cellcolor{heat!36}\textbf{63.7} & 63.7 & 0.0 & \cellcolor{heat!11}\textbf{19.9} & \cellcolor{heat!11}\textbf{19.2} & 0.95 \\
 & Required & \cellcolor{heat!35}61.6 & 26.0 & 35.6 & \cellcolor{heat!10}\textbf{18.5} & 13.0 & 5.5 & \cellcolor{heat!10}18.5 & 13.0 & 5.5 & \cellcolor{heat!7}12.3 & \cellcolor{heat!7}11.6 & 0.96 \\
\addlinespace[5pt]
\multirow{3}{*}{\shortstack[l]{Claude\\Haiku 5.5}} & None & \cellcolor{heat!12}21.2 & 21.2 & 0.0 & \multicolumn{3}{c}{\color{sub}---} & \cellcolor{heat!12}21.2 & 21.2 & 0.0 & \cellcolor{heat!9}15.8 & \cellcolor{heat!9}15.8 & 0.95 \\
 & Optional & \cellcolor{heat!32}56.2 & 24.0 & 32.2 & \cellcolor{heat!21}\textbf{37.0} & 24.7 & 12.3 & \cellcolor{heat!21}\textbf{37.7} & 28.8 & 8.9 & \cellcolor{heat!18}\textbf{31.5} & \cellcolor{heat!17}\textbf{30.1} & 0.98 \\
 & Required & \cellcolor{heat!37}\textbf{65.8} & 17.1 & 48.6 & \cellcolor{heat!17}30.8 & 17.8 & 13.0 & \cellcolor{heat!17}30.8 & 17.8 & 13.0 & \cellcolor{heat!13}23.3 & \cellcolor{heat!10}17.8 & 0.97 \\
\midrule
\multirow{3}{*}{\textbf{Pooled}} & None & \cellcolor{heat!28}50.2 & 50.2 & 0.0 & \multicolumn{3}{c}{\color{sub}---} & \cellcolor{heat!28}50.2 & 50.2 & 0.0 & \cellcolor{heat!10}18.3 & \cellcolor{heat!9}16.2 & 0.95 \\
 & Optional & \cellcolor{heat!35}62.8 & 50.9 & 11.9 & \cellcolor{heat!7}12.8 & 8.7 & 4.1 & \cellcolor{heat!31}\textbf{54.8} & 51.8 & 3.0 & \cellcolor{heat!15}\textbf{25.8} & \cellcolor{heat!14}\textbf{24.2} & 0.96 \\
 & Required & \cellcolor{heat!39}\textbf{68.7} & 25.3 & 43.4 & \cellcolor{heat!14}\textbf{24.4} & 16.4 & 8.0 & \cellcolor{heat!14}24.4 & 16.4 & 8.0 & \cellcolor{heat!9}15.8 & \cellcolor{heat!7}13.2 & 0.98 \\
\bottomrule
\end{tabularx}
\endgroup
\caption{Pair Recall and conflict detection under common generation allowances. All entries except Cost are percentages. Two runs per model yield $n=146$ conflict observations per model/policy ($n=438$ per pooled policy). Total and Both give Pair Recall; at the first three stages, Total is divided into Same and Split (gray) by reader or contributor (Section~\ref{sec:metrics}). A dash denotes no worker channel. Cost is mean generation spending divided by the common \$59.60 per-round allowance, excluding judging. Cell shading is proportional to the value on a common 0--80\% scale. Bold marks the highest Pair Recall and Judge Pass rate within each model or the pooled rows.}
\label{tab:main-results}
\end{table*}

%\caption{Pair Recall and conflict detection under common generation allowances. Two runs per model yield $n=146$ conflict observations per model/policy ($n=438$ per pooled policy). Total/Both gives Pair Recall, divided into Same/Split by reader or contributor at the first three stages (Section~\ref{sec:metrics}). Here, --- denotes no worker channel. Cost is mean generation spending divided by the common \$59.60 per-round allowance, excluding judging. Bold marks the highest Total/Both and Judge Pass rates within each model or pooled Total.}
%\label{tab:main-results}
%\end{table*}

For conflict observation $i$, let $G_i=\{A_i,B_i\}$ be its planted pair and let $E_i^t$ contain the endpoints observed at stage $t$, pooled across eligible agents or artifacts.
Adapting FEVER's complete-evidence-group evaluation \citep{thorne2018fever,feverscorer}, we define \emph{Pair Recall} as
\begin{equation}
R_{\mathrm{pair}}^t=\frac{1}{N}\sum_{i=1}^{N}
\mathbf{1}\!\left[G_i\subseteq E_i^t\right].
\label{eq:pair-recall}
\end{equation}
A conflict earns pair credit when both endpoints are present, regardless of unrelated evidence.
Table~\ref{tab:main-results} labels this quantity Total at the first three stages and Both at Final Report.

At the first three stages, we separate \emph{Same}, where one individual supplies both endpoints, from \emph{Split}, where only the contributors' union contains both.
Same takes precedence when multiple routes exist, giving $R_{\mathrm{pair}}^t=R_{\mathrm{same}}^t+R_{\mathrm{split}}^t$.
These labels describe cumulative provenance across the trajectory, identifying readers at Source Reads, sending workers at Worker Reports, and immediate contributors at Lead Evidence.
For example, under Optional, a lead-read $A_i$ and worker-supplied $B_i$ form Split, whereas one worker supplying both forms Same.
Because Final Report is a single artifact, it uses Both without a Same/Split breakdown.

An LLM judge (\texttt{claude-sonnet-5}) grades the reports, using model-based evaluation \citep{zheng2023judge}.
Our rubric has two criteria per conflict: identifying the discrepancy and accurately treating its quantitative comparison.
\emph{Judge Pass} requires both criteria, including an explicit flag for the specified conflict; its rate measures conflict detection.
We evaluate semantic grades alongside endpoint presence.
Judge Pass evaluates each planted conflict individually.

\section{Experimental Results}
\label{sec:results}

\begingroup
\brokenpenalty=10000
We first examine models' delegation choices and worker contributions (\S\ref{sec:delegation-behavior}).
We then compare evidence collection with conflict detection and trace where Pair Recall declines (\S\ref{sec:conversion}).
Finally, we compare Required and None on matched cases grouped by whether one worker reads both conflicting claims or the claims are split across workers (\S\ref{sec:matched}).
\par\endgroup

\subsection{Experimental Setup}
\label{sec:setup}
% Queue Figure 4 on page 6 so that it opens page 7, after Table 1.
% Edit figures/source/sankeys_tidy.svg and run scripts/export_figures.py 4.
\begin{figure*}[!t]
  \centering
  \includegraphics[width=0.95\textwidth]{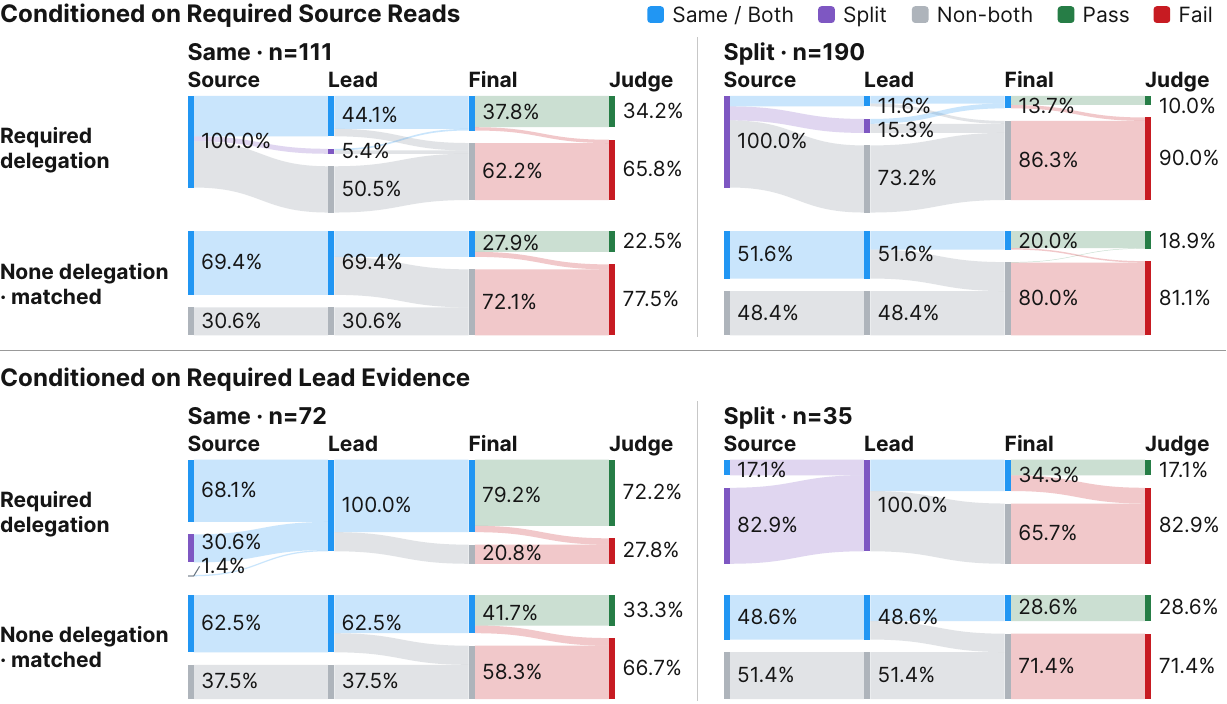}
  \caption{Examination of how multi-agent memory consistency relates to conflict detection
  via conditioned Sankey diagrams comparing matched cases.
  Pooled across runs, Required has higher Judge Pass than matched None in
  the Same cohorts, but lower Judge Pass in the Split cohorts. This pattern
  holds for both Source Reads and Lead Evidence groupings.}
  \label{fig:pooled-matched}
  \label{fig:pooled-source}
  \label{fig:pooled-lead}
\end{figure*}

% Keep at least two ending lines together across a column break.
\begingroup
\widowpenalty=10000
We evaluate GPT-5.6 Luna and GPT-6 Luna through the Codex SDK, and Claude Haiku 5.5 through the Claude Code SDK.
Each model runs each policy twice on all 11 tasks, yielding 198 task executions and 1,314 conflict observations.
This gives 146 observations per model/policy and 438 per pooled policy, with repeated conflicts clustered within tasks.
\par\endgroup

% Avoid splitting a hyphenated word across the page beneath the floats.
\begingroup
\brokenpenalty=10000
Generation uses common per-task dollar allowances totaling \$59.60 per round.
When spending remains below a configured fraction of the allowance, the system continues its native session to review and revise the report, retaining its history and cumulative spending.
Accounting covers all lead and worker calls, including revisions, compaction, and paid interruptions; judging is excluded.
Because a phase finishes before the controller checks whether to stop, spending can exceed the target budget, with mean spending per model/policy ranging from 0.95 to 1.02 times the allowance.
\par\endgroup

Table~\ref{tab:main-results} reports Pair Recall, evidence provenance, Judge Pass, and spending for each model/policy.
Figure~\ref{fig:study-insights} summarizes the main patterns in these results.

\subsection{Delegation Behavior and Worker Contributions}
\label{sec:delegation-behavior}

Under Optional, Haiku creates workers in all 22 executions, whereas GPT-5.6 does so in 9/22 and GPT-6 in 10/22 (Figure~\ref{fig:study-insights}a).
Workers also contribute different amounts of evidence.
Across all 146 conflict observations per model, both GPT models' Worker Reports achieve Pair Recall of only 0.7\% (1/146) under Optional, compared with 24.0\% under Required for GPT-5.6 and 18.5\% for GPT-6.
Haiku's workers achieve 37.0\% under Optional and 30.8\% under Required.

Required lets us compare evidence preservation under mandatory delegation across models with different tendencies to delegate.

\subsection{From Evidence Collection to Conflict Detection}
\label{sec:conversion}

Pooled across models and runs, Required increases Source Reads Pair Recall to 68.7\%, compared with 50.2\% under None, but Judge Pass falls from 16.2\% to 13.2\% (Figure~\ref{fig:study-insights}b).
Optional reaches 62.8\% Source Pair Recall and 24.2\% detection.
Collecting more evidence at the source stage therefore does not guarantee higher conflict detection in the final report.

The gap is visible within each model.
Relative to None, Required raises Source Reads Pair Recall by 8.9 percentage points (pp) for GPT-5.6 and 2.1~pp for GPT-6, while detection falls by 5.5~pp for each.
Haiku gains 44.5~pp in Source Pair Recall but only 2.1~pp in detection.
Optional has the highest detection rate for all three models.

Figure~\ref{fig:study-insights}c shows the difference between Source Reads and Final Report Pair Recall, divided into the decline before and after Lead Evidence.
Required has a larger total decline than None for every model, with most of its decline occurring before the lead.
Under None, source reads enter the lead's evidence, so there is no drop between those two stages.
Each bar measures percentage-point changes over all 146 conflict observations for that model and policy.

\begingroup
\widowpenalty=10000
Across all 438 Required observations, Source Reads contain both claims in 301 cases (68.7\%), compared with 107 (24.4\%) at Worker Reports and Lead Evidence.
Of the 301 pairs present at Source Reads, 106 also appear in Worker Reports (35.2\%).
Worker Reports and Lead Evidence have identical labels, locating the observed handoff gap between source inspection and worker reporting.
Collective retrieval therefore often fails to provide the lead with both claims needed to detect a conflict.
\par\endgroup

\subsection{Evidence Distribution in Matched Cases}
\label{sec:matched}

\begingroup
\widowpenalty=10000
Figure~\ref{fig:pooled-matched} examines whether conflict detection differs when one agent supplies both claims and when the claims are split across agents.
Same and Split cohorts are defined by Required's observed evidence distribution, with None evaluated on the identical model, run, task, and conflict instances.
The upper group conditions on Source Reads and the lower group on Lead Evidence.
Each cohort's size $n$ applies to both policies, with percentages calculated within that cohort.
\par\endgroup

Pooled across runs, Required achieves Judge Pass on 38/111 observations (34.2\%) in the Source Reads Same group, compared with 25/111 (22.5\%) under None.
In the Split group, it detects 19/190 (10.0\%), compared with 36/190 (18.9\%).
Conditioning on Lead Evidence gives the same direction: 72.2\% versus 33.3\% for Same, and 17.1\% versus 28.6\% for Split.
Required's advantage over None is concentrated in cases where one agent supplies both claims.

Shared-draft reads and arithmetic reconstruction can change a pair from Split at Source Reads to Same at Lead Evidence.
The two groupings therefore describe evidence contributors at their respective stages.

\FloatBarrier
\section{Conclusion}
\label{sec:conclusion}

%Our benchmark and profiler provide a direct testbed for interventions that preserve cross-assignment evidence during worker reporting. Based on our analysis, \emph{Required} broadens evidence collection, but workers report both claims for only 35.2\% of retrieved pairs. This leaves a substantial gap at the worker-to-lead handoff.
%In pooled matched comparisons, \emph{Required} detects more conflicts than single-agent execution when one worker reads both claims, but fewer when the claims are split across workers.

Answering where factual conflicts disappear in multi-agent LLM systems requires separating retrieval coverage from distributed memory consistency.
While mandatory delegation (Required) broadens collective source retrieval over single-agent execution (None), net conflict detection drops because evidence fractures across two stages.
First, at the worker-to-lead handoff, workers summarizing isolated partitions lack the cross-document context needed to recognize figures that appear unproblematic in isolation as conflicting across documents, omitting one or both claims in 73.2\% of split-evidence cases compared with 50.5\% when a single worker reads both.
Second, even when both conflicting claims reach the lead, conflict resolution depends heavily on provenance: leads flag 72.2\% of conflicts supplied by a single contributor (Same), yet only 17.1\% when claims arrive divided across separate worker reports (Split).

For system designers, these findings caution against rigid, non-overlapping context partitioning on cross-document verification tasks; permitting overlapping source access and lead source inspection (Optional) consistently outperforms both single-agent and strictly partitioned execution.
More broadly, end-to-end task accuracy masks opposing retrieval gains and coordination losses.
Our benchmark and evidence-flow profiler allow practitioners to separate source discovery, handoff retention, and lead synthesis, providing a testbed for structured reporting protocols and conflict-preserving multi-agent memory.

\section*{Limitations}

Our evidence comes from 11 Harvey LAB tasks with planted numerical conflicts and two runs per model.
Broader document domains, non-numerical conflicts, and more repetitions would test how widely the observed patterns hold, with uncertainty estimates accounting for shared tasks.
The profiler measures cumulative evidence presence; tracing when both claims become jointly available would extend this analysis.
Manual inspection identified three saved Judge Pass labels that credited neighboring comparisons rather than the planted conflict; broader human review could refine endpoint matching and semantic grading.

Required combines mandatory delegation, partitioned source assignments, and restricted lead access, while model families use different SDKs.
The matched groups also reflect agents' observed retrieval and reporting choices.
Supplying identical evidence while varying which agents receive it would help separate the effect of distribution from retrieval difficulty.
Reporting interventions could then test whether preserving paired claims improves detection.
Budgets were calibrated from observed spending, and GPT-5.6 Required checkpoints were selected retrospectively; future evaluations should fix budgets and checkpoint rules in advance.

\FloatBarrier
\bibliography{paper}
\end{document}